\documentclass{llncs}
\usepackage{svg}
\usepackage{biblatex}
\usepackage{amsmath}
\usepackage{longtable}
\newcommand*{\qm}{\texttt{"}}
\bibliography{bib.bib}
\begin{document}
\title{Deep learning model solves change point detection for multiple change types}
\titlerunning{CPD}  
%
\author{Alexander Stepikin\inst{1, 2, 3}, Evgenia Romanenkova\inst{2}, Alexey Zaytsev\inst{2} \\ 
\email{stepikin.al@phystech.edu} \\ }
\authorrunning{Alexander Stepikin et al.} 
\institute{Institute for Information and Transmission Problems, Moscow\\
\and
Skolkovo Institute of Science and Technology, Moscow\\
\and
Moscow Institute of Physics and Technology, Moscow}

\maketitle              

\begin{abstract}
A change points detection aims to catch an abrupt disorder in data distribution. Common approaches assume that there are only two fixed distributions for data: one before and another after a change point. Real-world data are richer than this assumption. There can be multiple different distributions before and after a change. We propose an approach that works in the multiple-distributions scenario.
Our approach learn representations for semi-structured data suitable for change point detection, while a common classifiers-based approach fails. Moreover, our model is more robust, when predicting change points. The datasets used for benchmarking are sequences of images with and without change points in them.
\keywords{change point detection, representation learning, sequential data}
\end{abstract}
\section{Introduction}
The modern world is full of time series and sequential data: from server logs and sensors' reading while drilling to video analysis \cite{aminikhanghahi2017survey, romanenkova2021principled, romanenkova2019real}. The change point is an abrupt alteration of the distribution the data comes from. Its occurrence means that the generating process has dramatically changed what may be an emergency signal. Thus, there is an urgent need to detect those shifts in a fast and accurate way.

The basic approaches mostly assume that there are only two unknown but fixed data distributions: one before and one after a change point. On the contrary, the real-world tasks require a method suitable for the case of multiple possible distributions before and after change point, for example, in a video monitoring system \cite{romanenkova2021principled}:
we observe numerous types of system behaviour and are interested in their changes.

We adapt the existing method of change-point detection based on the representation learning to the new setting with multiple possible distributions. 

We compare the performance of the recurrent neural network with principled loss function to the one with basic binary cross-entropy loss in terms of the key change point detection and classification metrics of quality. The results show that our model can make precise and confident predictions in different experimental setups with up to $10$ different types of distribution changes.
The considered changes are in Table~\ref{table:changes}.

\begin{table}[h!]
    \centering
    \begin{tabular}{ll}
    \hline
    Change number & New considered \\
      & changes \\
    \hline
    $1$ & $[4,7]$ \\
    $2$ & $[7,4]$ \\
    $4$ & $[1,9]$, $[9,1]$\\
    $6$ & $[2,5]$, $[5,2]$\\
    $8$ & $[0,8]$, $[8,0]$\\
    $10$ & $[3,6]$, $[6,3]$\\
    \hline
    \end{tabular}
    \caption{Considered types of changes for different number of considered changes. $[x,y]$ designates one change with $x$ being a digit before a change and $y$ being a digit after a change}
    \label{table:changes}
\end{table}

\vspace{-1.5cm}
\section{Related work}
The statistical approach includes the use of well-known statistics such as cumulative sum, Shiryaev-Roberts statistics, posterior probability statistics, or the use of smoothing and various filters, for example, the Kalman filter, as well as various combinations of them \cite{yang2006adaptive, burn10, aminikhanghahi2017survey}. Such statistics find application in various fields because of their simplicity but strong performance. However, they suggest the fixed distribution before and after a change point. So, their performance may suffer in the case of multiple distributions. 

The seminal book \cite{shiryaev2017stochastic} introduces an accurate overview of the mathematical background behind these approaches and presents several examples of change point detection applications in financial analysis. In the article \cite{burn10}, the authors use ensembles of \qm weak\qm detectors (classical methods, but operating under conditions of assumptions violation) to predict the degradation of systems with extensive software in real-time. Paper \cite{yang2006adaptive} provides another example of the usage of such approaches for medical issues. The authors combine Kalman filters, cumulative sums, and an exponentially weighted moving average to detect changes in the heart rate trend in real-time. 

Statistical approaches were developed in the paper \cite{romanenkova2021principled}. Based on criteria from \cite{shiryaev2017stochastic}, the authors present differentiable loss functions for change point detection via neural network. In this work, we consider this approach, so more details are presented in the section \ref{sec:methods}.

Accurate reviews of state-of-the-art change point detection methods can be found in \cite{aminikhanghahi2017survey, truong2020selective}. Authors describe different categories of methods such as methods based on probability density ratio (both parametric and non-parametric), subspace models, probabilistic models, kernel-based methods, clustering, graph-based methods and classification methods. 

\section{Methods}\label{sec:methods}
In this section, we introduce the problem of change point detection (CPD), the investigated approach based on the principled loss function presented in the paper \cite{romanenkova2021principled} and the design of the experiments. 
\subsection{Problem statement}
The basic formulation of the change point detection problem is the following. Consider a random process $X^{1:T} = \{x_{1},\ldots,x_{T}\}$ of length $T\leq \infty$ consisting of observations $x_{t}\in \bbbr^{d}$, $t \in \{1, ..., T\}$. The process depends on a latent random variable $\theta\in\{0,1\ldots,+\infty\}$ that comes from an unknown distribution $G$. For each moment $t<\theta$, we assume the \qm normal\qm\ behaviour of the process $\ x_{t}\sim f_{\infty}$, and for $t\geq\theta$, it is \qm abnormal\qm\ $\ x_{t}\sim f_{0}$. The distributions $f_{0}$ and $f_{\infty}$ are unknown as well. The problem is to detect the true change point $\theta$ as quickly as possible.

It should be emphasized that we work in an \qm online\qm\ setup which means that the decision about a presence of a change point is based on the information available at a current moment of time.
\subsection{Principled loss function}

The main idea of the method is to train a neural network parameterized by the vector of weights $w$ with a special loss function \cite{romanenkova2021principled}. In other words, the model's input is a dataset $D = \{(X_{1}, \theta_{1}),\ldots,(X_{N},\theta_{N})\}$ , where $X_{i}$ denotes the $i$-th process and $\theta_{i}$ is the corresponding change point. The model $f_{w}$ produces outputs $p_{i} = \{p_{t,i}\}_{t = 0}^{T}$ where $p_{t,i} = f_{w}(X_{i}^{1:t})$ is the probability of the change point in the process $X_{i}$ at the particular time moment $t$.

The proposed loss function \eqref{eq:1} is a sum of two terms that, as shown in~\cite{romanenkova2021principled}, approximate the expected detection delay~\eqref{eq:2} and the expected time to false alarm if it appears~\eqref{eq:3} we aim to optimize, according to~\cite{shiryaev2017stochastic}:
\begin{equation}
    \label{eq:1}
    \mathcal{L}(f_{w}, D) = \mathcal{L}_{delay}(f_{w}, D) + c\cdot\mathcal{L}_{FA}(f_{w}, D),
\end{equation}
where

\begin{equation}\label{eq:2}
    \mathcal{L}_{delay}(f_{w}, D) = \frac{1}{N}\sum_{i=1}^{N}\left(\sum_{t=\theta_{i}}^{T}{(t-\theta_{i})p_{t,i}\prod_{k=\theta_{i}}^{t-1}(1 - p_{k,i}) +  (T+1-\theta_{i})\prod_{k=\theta_{i}}^{T}(1-p_{k,i})} \right)
\end{equation}

\begin{equation}\label{eq:3}
    \mathcal{L}_{FA}(f_{w}, D) = - \frac{1}{N}\sum_{i = 1}^N \left( \sum_{t = 0}^{\tilde{T}} t p_{t, i} \prod_{k = 0}^{t - 1} (1 - p_{k,i}) - (\tilde{T} + 1) \prod_{k = 0}^{\tilde{T}} (1 - p_{k,i}) \right),
\end{equation}
$\tilde{T} = \min{(\theta_i, T)}$. 

\subsection{Final pipeline}
\begin{figure}
    \centering
    \includegraphics[width=\linewidth]{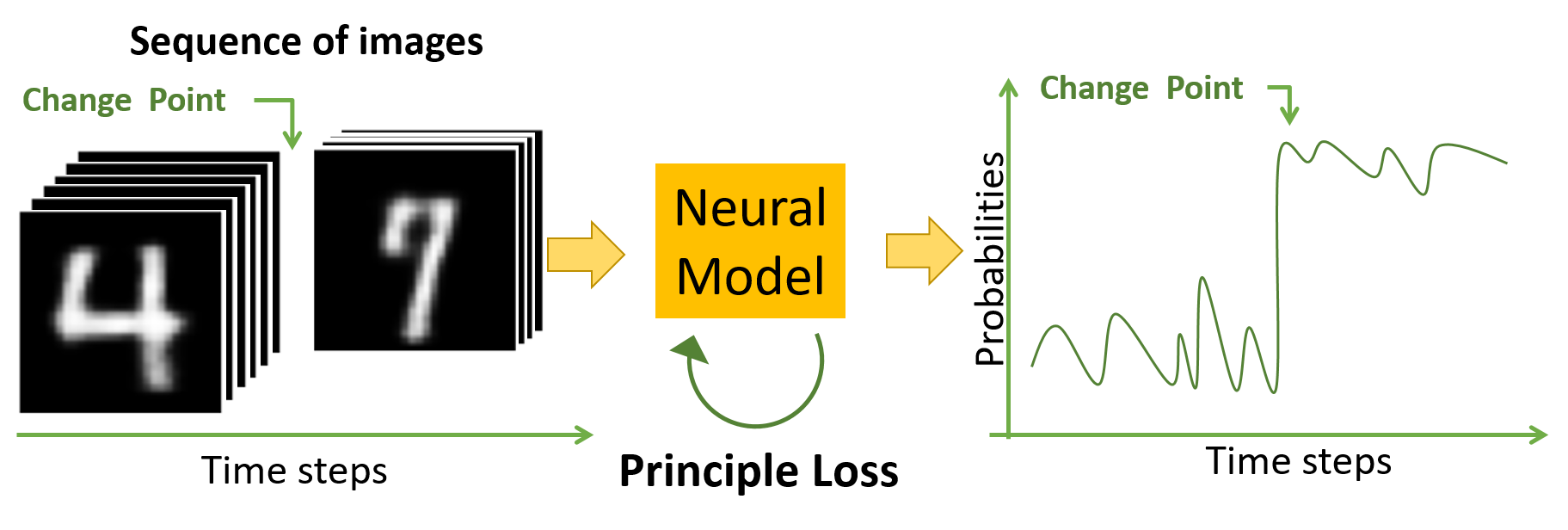}
    \vspace{-1cm}
    \caption{Teaser of investigated approach. For each moment, the model outputs the probability of change point.}
    \label{fig:teaser}
\end{figure}

As a result, the following end-to-end algorithm has been proposed summarized in Figure~\ref{fig:teaser}.
Training of a neural network for change point detection consists of the following steps:
\begin{enumerate}
    \item Using a neural network $f_{w}$, get the change point probabilities $p_{i} = f_{w}(X_{i})$ for each input $X_{i}$ in the dataset;
    \item Calculate the loss function given by~\eqref{eq:1};
    \item Train the neural network via SGD.
\end{enumerate}

For making predictions, report about the change point if the probability $p_{t,i}$ exceeds a pre-selected threshold $s$.

\vspace{-1cm}
\subsection{Classification approach}

Alternatively, we can think of this problem as a binary classification: it is necessary to classify all the process observations, whether they occurred before or after the distribution change. In this case, it is possible to apply a deep model with standard binary cross-entropy (BCE) loss.
\subsection{Multiple types of change points}
In this paper, we investigate a more complicated scenario when there are multiple types of change points. In particular, it means that the observations after the change point may come from several fixed distributions: $f_{0,\infty}^{(1)}, f_{0,\infty}^{(2)}, \ldots f_{0,\infty}^{(K)}$. 

We state that our embeddings would be universal enough to deal with the case of multiple distributions. Our suggestion is that a proper selection of hyperparameters can work in this case. So, a careful adaptation of existing methods can make a neural network learn to detect the change itself even in case of various scenarios for changes. However, at the same time, theoretical guarantees exist only for the case of fixed distribution.

Thus, our paper focuses on the possibility to deal with the challenging multiple distributions cases.
%
\section{Results}
In this section, we explain the conducted experiments and present their results.


\subsection{Datasets}
%

Following \cite{romanenkova2021principled}, we used the datasets consisting of sequences of MNIST-like images with smooth transitions from digit $x$ to digit $y$. The sequences were generated by a Conditional Variational Autoencoder (CVAE). Each sequence has the length of $64$ images and comprises of $3$ segments: $[x,x]$, $[x,y]$ and $[y,y]$. The middle segment has a random length from $1$ to $10$ images. If $x\neq y$, the sequence has a change point.

In all the experiments, datasets were balanced, i.e. they included equal amounts of \qm normal\qm\ (without a change of digit) and \qm abnormal\qm\ (with a transition from one digit to another) sequences. We constructed different datasets: with $1$, $2$, $4$, $6$, $8$ and $10$ types of changes in data distributions. More detailed, for different types of changes, we used different starting and ending digits listed in Table~\ref{table:changes}. We generated $500$ sequences with each type of change point and $500$ corresponding sequences without a digit transition. As a result, we got $6$ balanced datasets consisting of $1000$, $2000$, $4000$, $6000$, $8000$ and $10000$ sequences respectively.

\begin{figure}[ht]
    \begin{minipage}{0.5\textwidth}
    \centering
    \includegraphics[width=\linewidth]{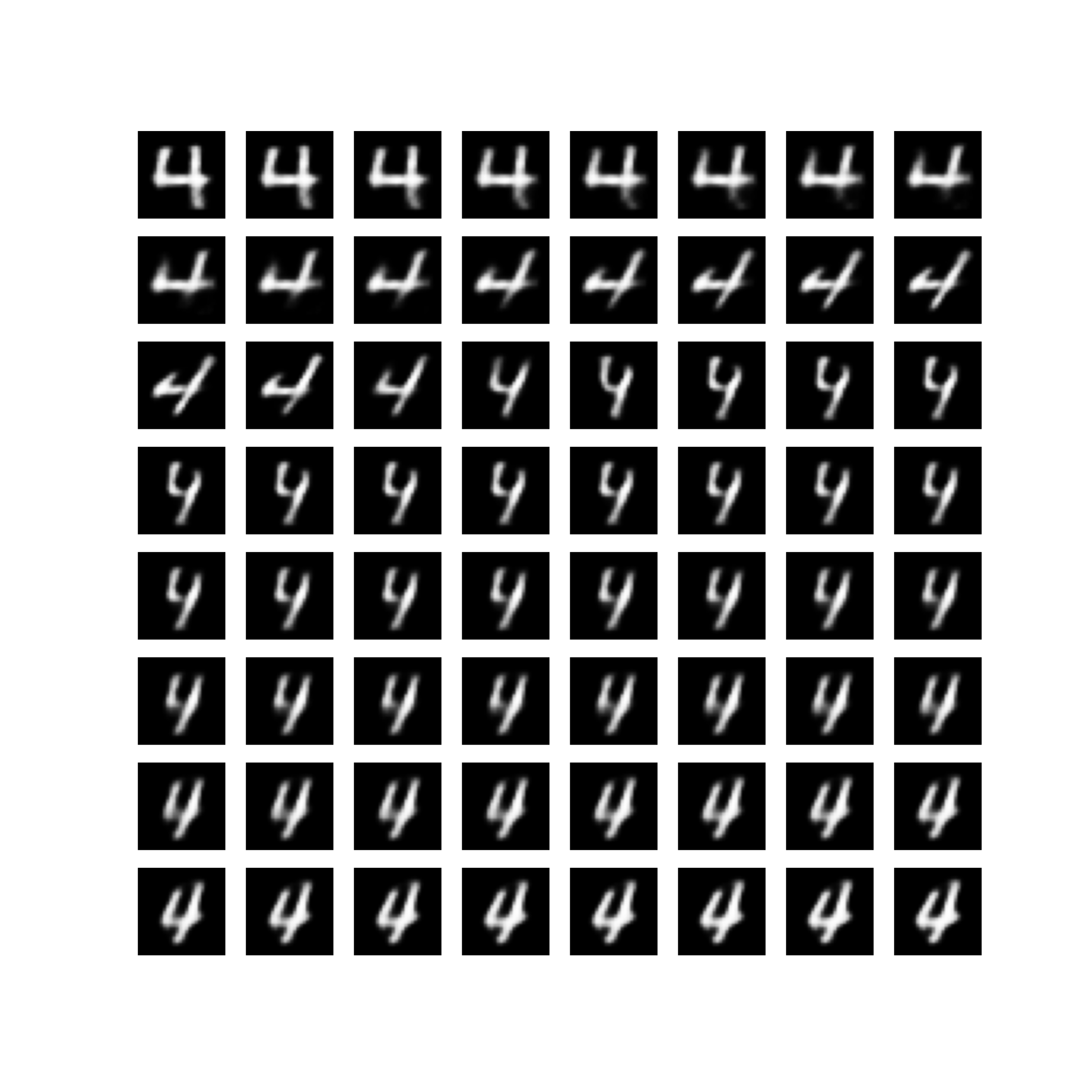}
    \end{minipage}\hfill
    \begin{minipage}{0.5\textwidth}
    \centering
    \includegraphics[width=\linewidth]{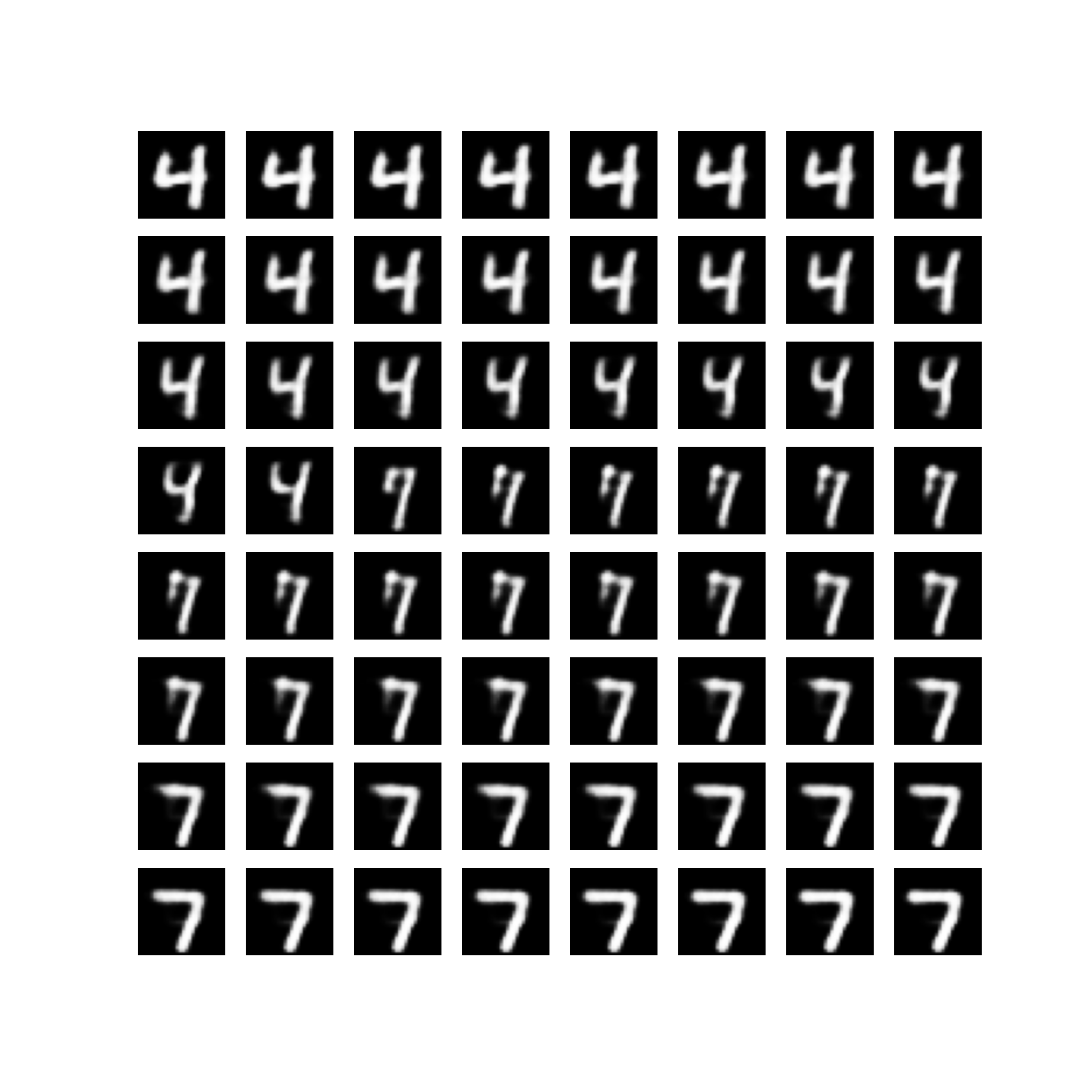}
    \end{minipage}
    \vspace{-1cm}   
    \caption{Examples of generated sequences with (left) and without (right) a transition of the digits.}
    \label{fig:sequences}
\end{figure}

The examples are shown in Figure~\ref{fig:sequences}.

\subsection{Models architecture}
We used models with the following architecture: recurrent neural network (RNN) followed by $2$ fully-connected layers with ReLU activations and the output layer with sigmoid activation. We chose LSTM \cite{hochreiter1997lstm} as a basic model because it handles the sequential nature of the data. The code of data generation and the experiments is available online\footnote{GitHub: \url{https://github.com/romanenkova95/PrincipledCPD}}.
\subsection{Comparison details}

In all the experiments, we compare the performance of two deep models with the same architecture but different loss functions: custom CPD loss~\eqref{eq:1} with balanced weights of the terms and BCE loss. All the other parameters were identical. For the model with CPD loss, we tried different hyperparameters $T$. 
\subsection{Main metrics}
In order to evaluate the models` performance, we used different metrics: standard classification metrics and specific metrics for change point detection.
\paragraph{Classification metrics.}
We interpret classification metrics for the change point detection problem in the following way.
\begin{enumerate}
    \item True Positive (TP): the change point was correctly detected in the \qm abnormal\qm\ sequence no earlier than it appears.
    \item False Positive (FP): a model predicts a change point in a \qm normal\qm\ sequence or reports about it before it appears in an \qm abnormal\qm\ one.
    \item True Negative (TN): a model does not predict a change point in a \qm normal\qm\ sequence.
    \item False Negative (FN): a model does not report about a change point in a sequence where there is one.
\end{enumerate}
Having these elements of the confusion matrix, we calculate accuracy, recall and $F_{1}$-score based on the predictions of different models.
\paragraph{Covering.}
As a specific metric for change point detection, we adopted the Covering metric presented in \cite{van2020evaluation} that comes from the image segmentation problem and evaluates the similarity of two partitions of the sequences: the one based on the model`s predictions and the true one. The higher is the covering metric, the better is the performance.
\paragraph{Mean Detection Delay and Mean Time to False Alarm.}
These are the metrics commonly used for the evaluation of change point detection algorithms. Detection Delay $(\tau - \theta)^{+}$ is a difference between the change point time $\tau$ detected by a model and the real one $\theta$. Time to false alarm (FA) is equal to $\tau$  -- the difference between the first positive prediction if there is one and the beginning of the sequence. The custom loss function (\ref{eq:1}) is aimed to minimize the expected Detection Delay while keeping the expected time to FA large enough. 
\paragraph{Area under the detection curve (AUC).}
The detection curve is a graph where the $x$-axis is the mean time to FA, and the $y$-axis is the mean Detection Delay. The examples of the curves are in Figure~\ref{fig:curves}. The points on the graph were obtained by varying the detection threshold $s$ in the range from $0$ to $1$. As the problem of the reasonable threshold choice is not straightforward, the idea is to consider all the possible variants and measure areas under the detection curves of different models. The lower this area is, the better a model performs. 
\subsection{Experimental results}
\begin{figure}[!ht]
    \centering
    \includegraphics[width=\linewidth]{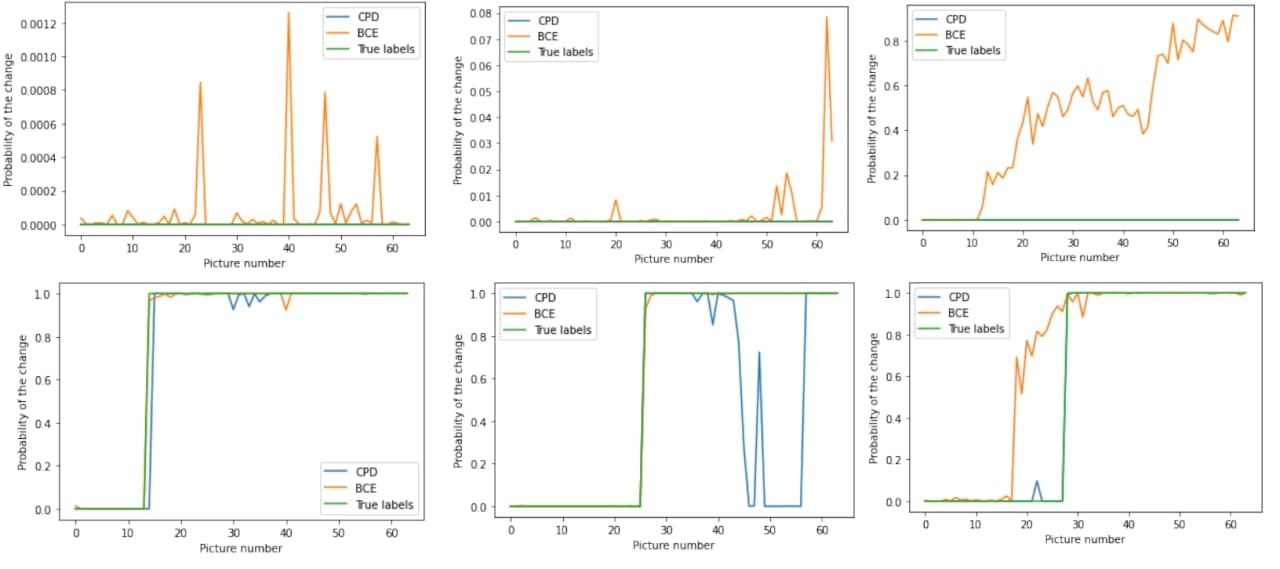}
    \vspace{-0.5cm}
    \caption{Typical predictions of the models in experiments with multiple types of distribution changes. The CPD model is more confident in her predictions, while for BCE loss we observe outliers in change prediction probabilities.}
    \label{fig:preds}
\end{figure}

\begin{figure}[!ht]
    \begin{minipage}{0.5\textwidth}
     \centering
     \includegraphics[width=\linewidth]{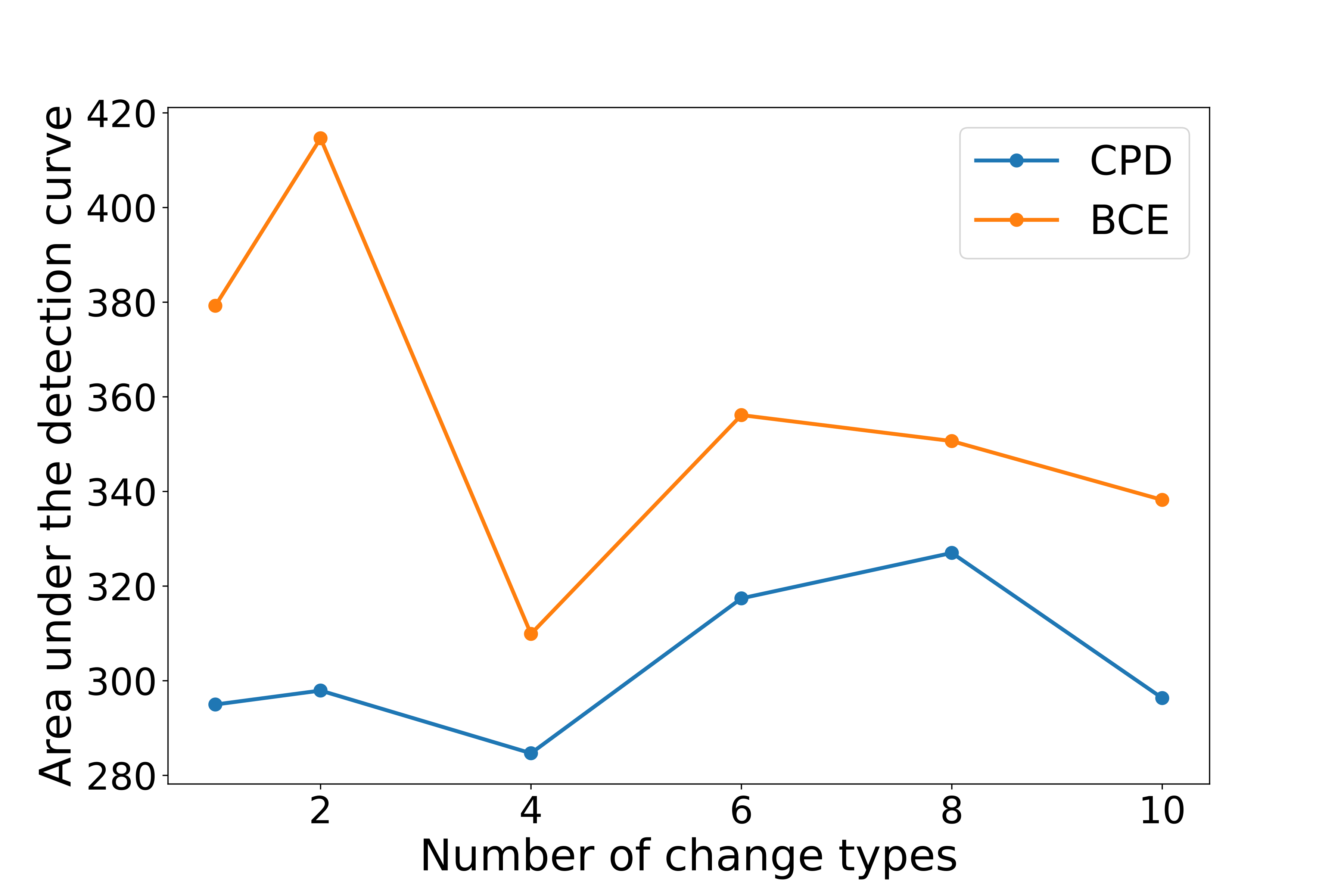}
   \end{minipage}\hfill
   \begin{minipage}{0.5\textwidth}
     \centering
     \includegraphics[width=\linewidth]{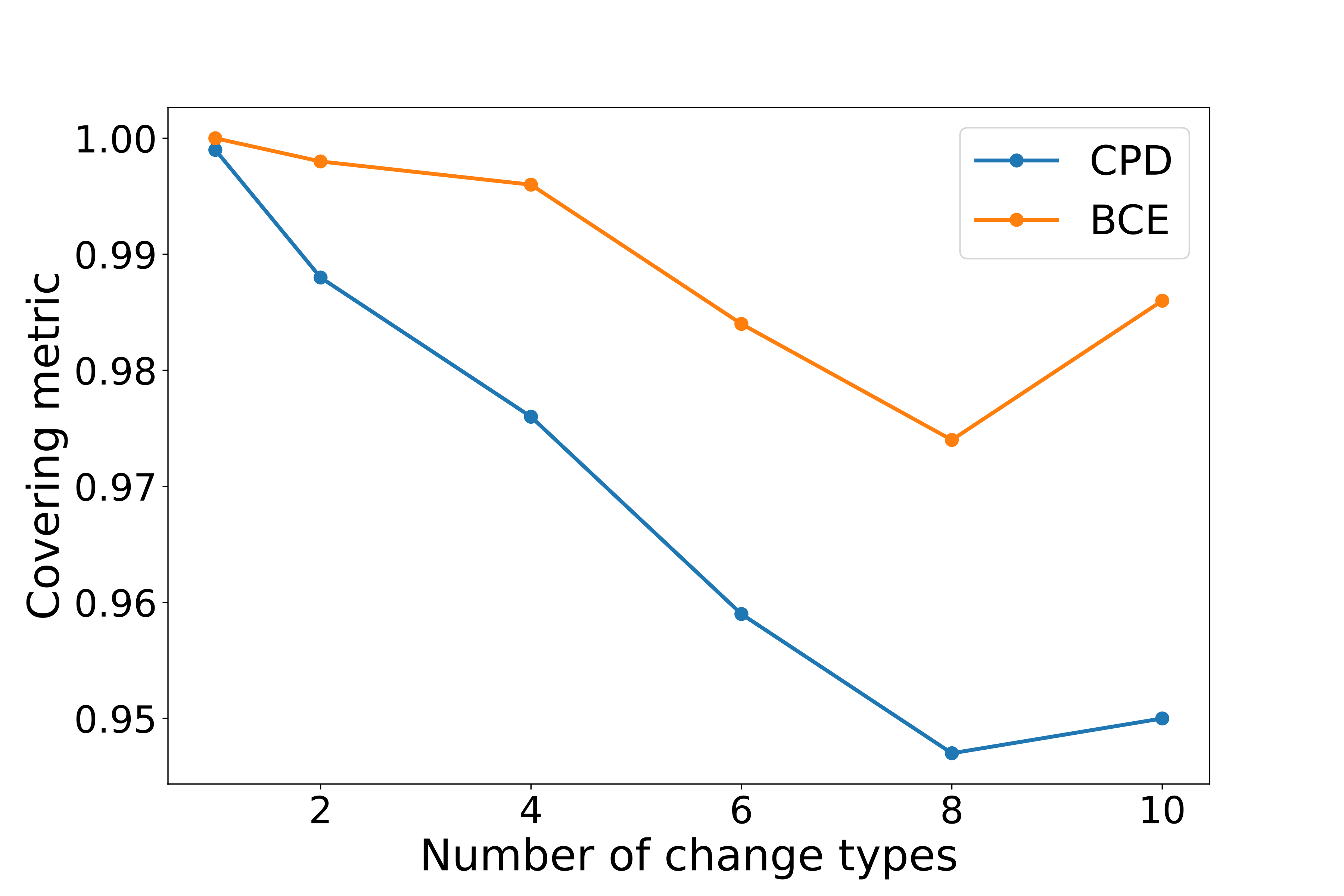}
   \end{minipage}
   \caption{Dependence of the main metrics (AUC $\downarrow$ and Covering $\uparrow$) on the number of change types. The quality of models` performance deteriorates as the complexity of the problem increases.}
   \label{fig:metrics}
\end{figure}

\begin{figure}[!ht]
   \begin{minipage}{0.5\textwidth}
     \centering
     \includegraphics[width=\columnwidth]{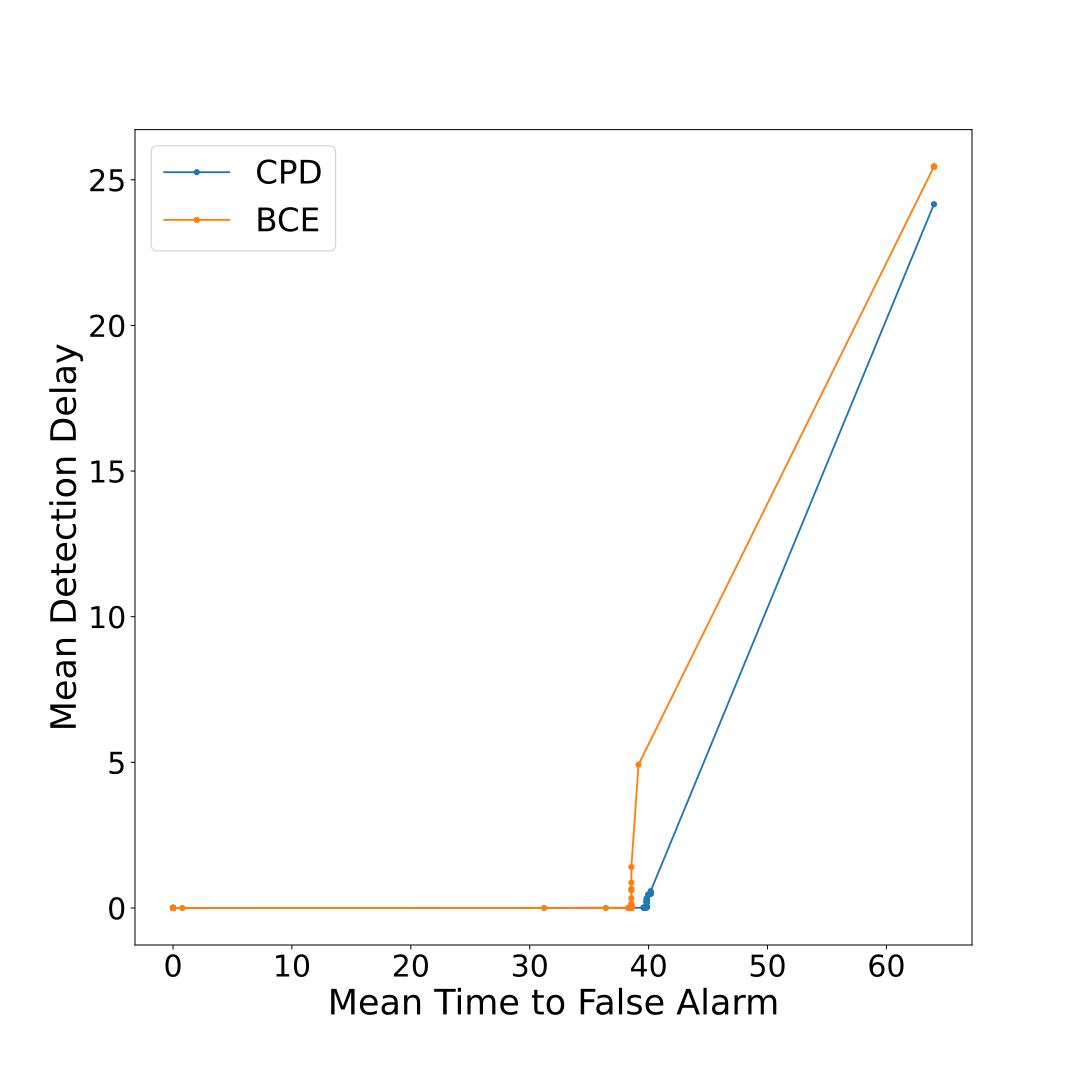}
     1 type of change
   \end{minipage}\hfill
   \begin{minipage}{0.5\textwidth}
     \centering
        \includegraphics[width=\columnwidth]{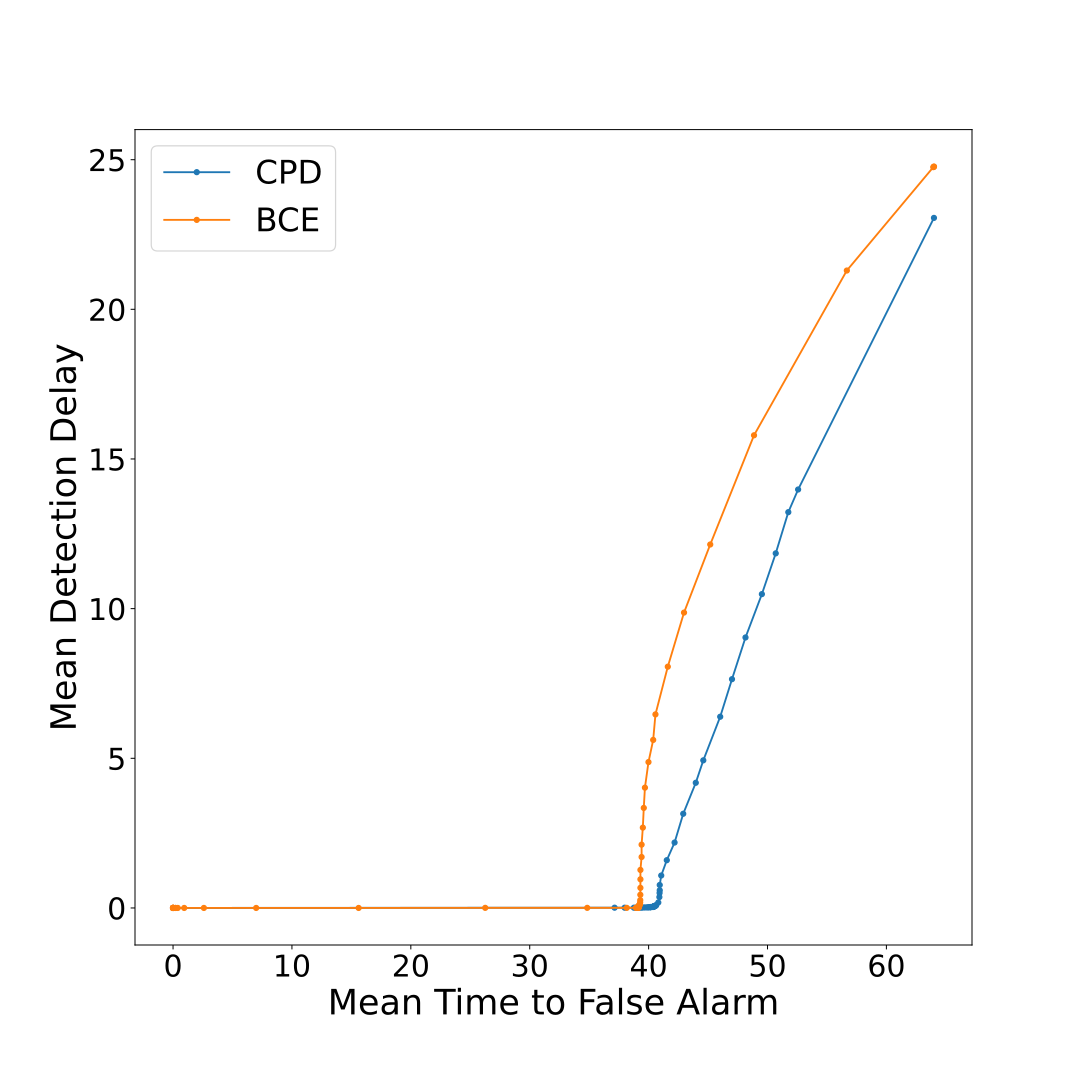}
        2 types of changes
   \end{minipage}\hfill
   \begin{minipage}{0.5\textwidth}
     \centering
        \includegraphics[width=\columnwidth]{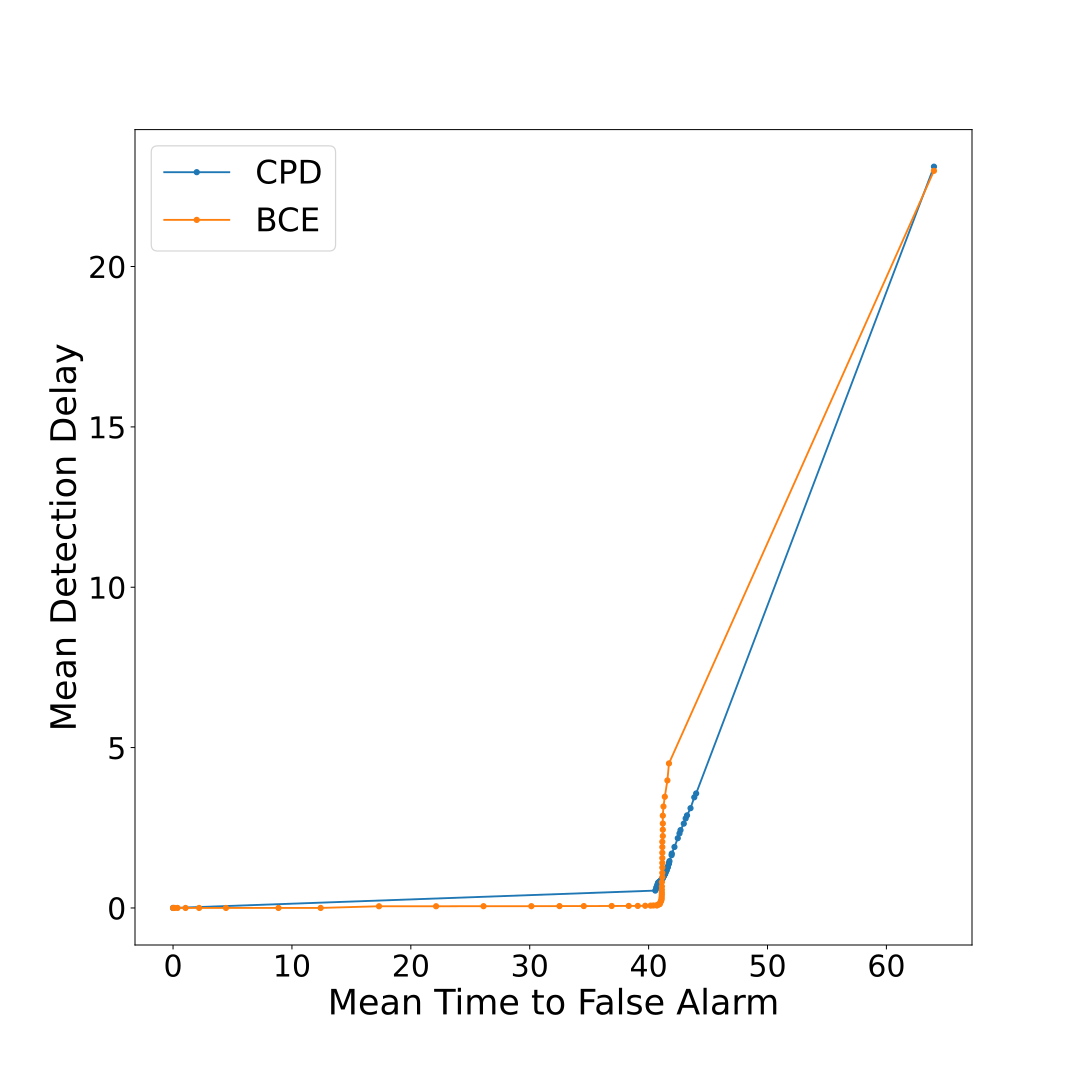}
        4 types of changes
   \end{minipage}\hfill
   \begin{minipage}{0.5\textwidth}
     \centering
        \includegraphics[width=\columnwidth]{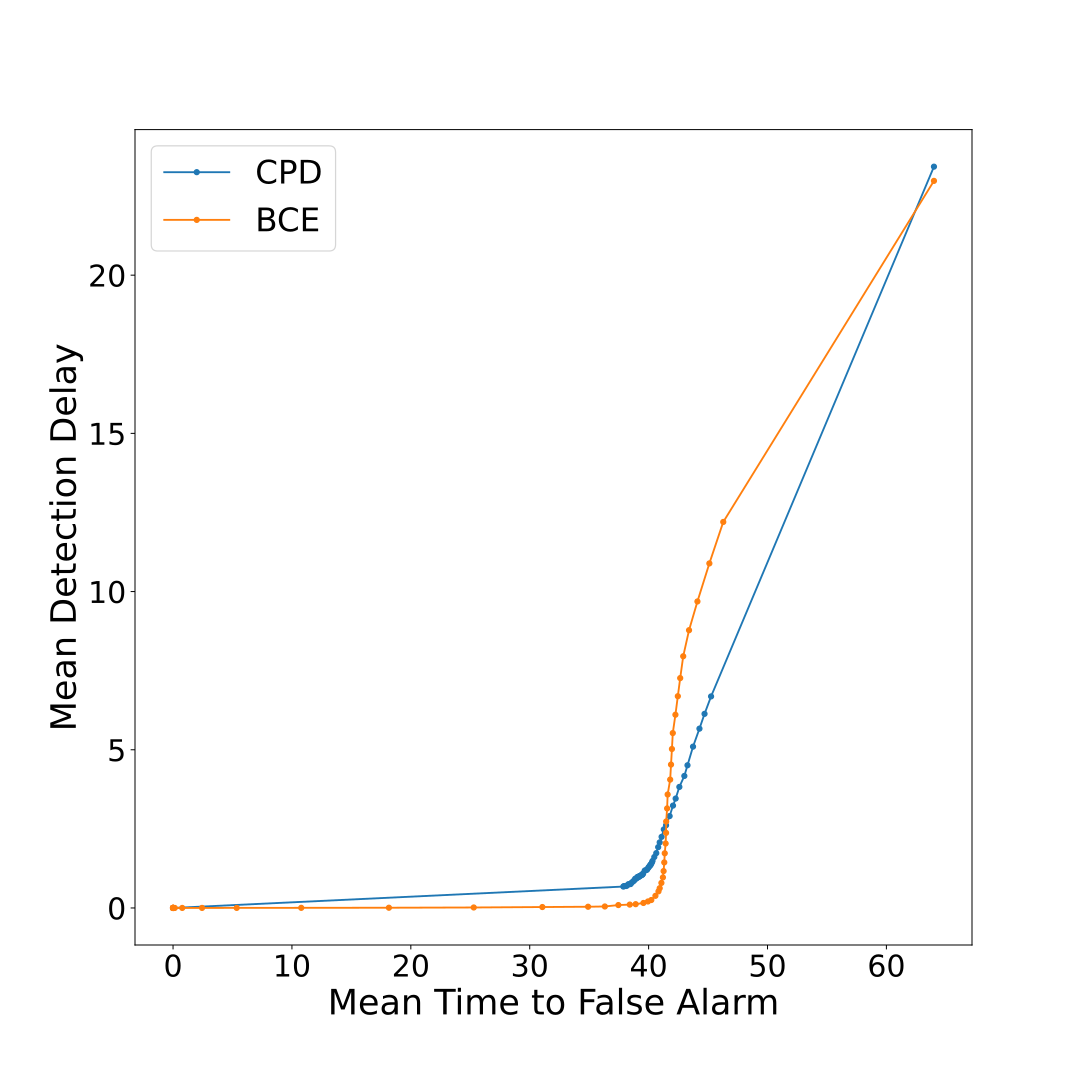}
        6 types of changes
   \end{minipage}\hfill
   \begin{minipage}{0.5\textwidth}
     \centering
        \includegraphics[width=\columnwidth]{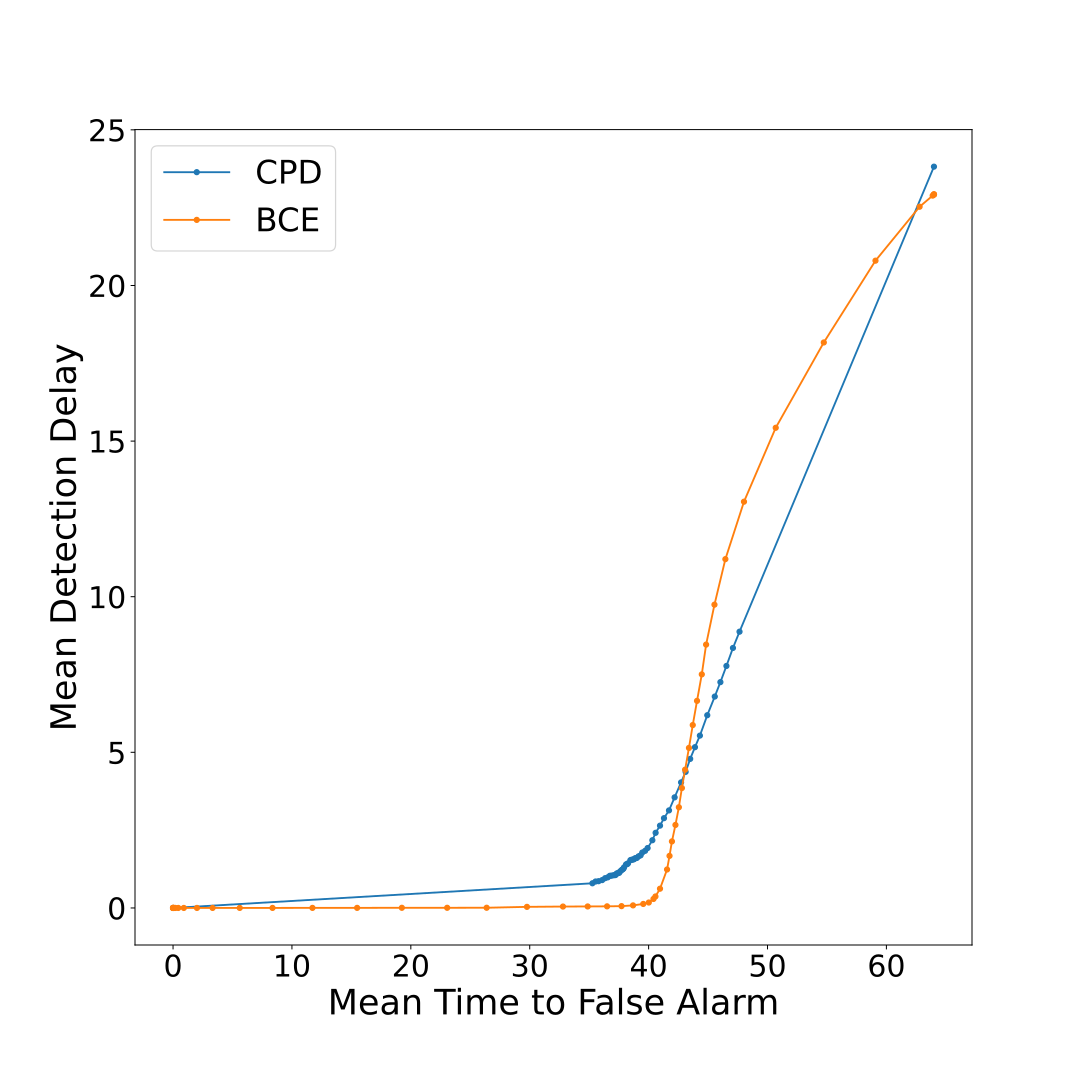}
        8 types of changes
   \end{minipage}\hfill
   \begin{minipage}{0.5\textwidth}
     \centering
        \includegraphics[width=\columnwidth]{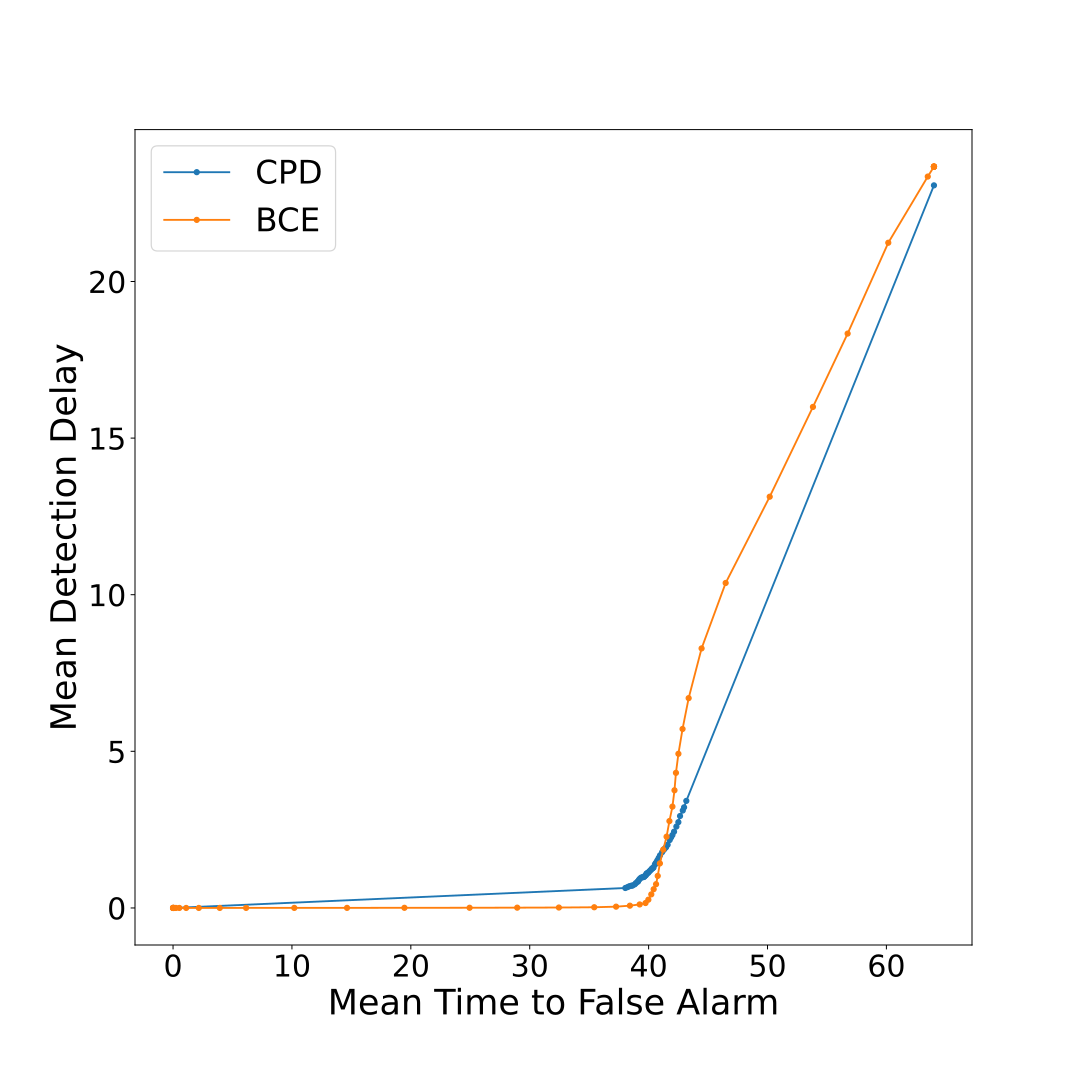}
        10 types of changes
   \end{minipage}
   \caption{Detection curves of the models with CPD and BCE losses for the MNIST experiments with multiple types of changes.}
   \label{fig:curves}
\end{figure}

We evaluated the main quality metrics with a fixed set of alarm thresholds in a range from $0.001$ to $0.999$. In addition, we calculated the areas under the detection curves based on the models` predictions. The results are presented in Table~\ref{table:metrcis} and Figure~\ref{fig:metrics}. For clarity, we chose to compare only those thresholds that optimize the considered metrics for different models. 
A few typical predictions of the both models in different experiments are depicted in the picture \ref{fig:preds}. The detection curves are depicted in the figure \ref{fig:curves}.

\begin{longtable}[!h]{c c c c c c c}
    \hline
        Loss & Threshold & Time to FA $\uparrow$ & Detection Delay $\downarrow$ & $F_1$-score $\uparrow$ & Covering $\uparrow$ & AUC $\downarrow$ \\ \hline
        &  &  & 1 type of change &  &  &  \\ \hline
        CPD & 0.01 & 39.84 & \textbf{0.03} & \textbf{1.00} & 0.998 & \textbf{294.98} \\
         & 0.1 & \textbf{41.64} & \textbf{0.03} & \textbf{1.00} & \textbf{0.999} &  \\ 
         & 0.2 & \textbf{41.64} & \textbf{0.03} & \textbf{1.00} & \textbf{0.999} &  \\ 
         & 0.5 & \textbf{41.64} & \textbf{0.03} & \textbf{1.00} & \textbf{0.999} &  \\ 
         & 0.7 & \textbf{41.64} & 0.05 & \textbf{1.00} & \textbf{0.999} &  \\ \hline
        BCE & 0.01 & 38.49 & \textbf{0.00} & 0.99 & 0.917 & 379.24 \\
         & 0.1 & \textbf{40.20} & \textbf{0.00} & \textbf{1.00} & \textbf{1.000} &  \\ 
         & 0.2 & \textbf{40.20} & \textbf{0.00} & \textbf{1.00} & \textbf{1.000} &  \\ 
         & 0.5 & \textbf{40.20} & \textbf{0.00} & \textbf{1.00} & \textbf{1.000} &  \\ 
         & 0.7 & \textbf{40.20} & 0.01 & \textbf{1.00} & \textbf{1.000} &  \\ \hline
         &  &  & 2 types of changes &  &  &  \\ \hline
        CPD & 0.001 & 39.63 & \textbf{0.05} & 0.89 & 0.844 & \textbf{297.93} \\
         & 0.01 & 39.74 & \textbf{0.05} & 0.91 & 0.949 &  \\
         & 0.5 & 40.19 & 0.09 & 0.93 & 0.985 &  \\ 
         & 0.99 & 41.50 & 0.19 & \textbf{0.97} & \textbf{0.988} &  \\
         & 0.9999 & \textbf{41.59} & 2.88 & 0.96 & 0.947 &  \\ \hline
        BCE & 0.001 & 8.59 & \textbf{0.00} & 0.16 & 0.459 & 414.69 \\
         & 0.01 & 39.15 & 0.01 & 0.89 & 0.847 &  \\ 
         & 0.5 & 40.56 & 0.02 & 0.96 & \textbf{0.998} &  \\ 
         & 0.99 & 39.36 & 0.88 & \textbf{1.00} & 0.980 &  \\ 
         & 0.9999 & \textbf{41.49} & 4.63 & 0.99 & 0.895 &  \\ \hline
         &  &  & 4 types of changes &  &  &  \\ \hline
        CPD & 0.001 & 41.76 & \textbf{0.70} & 0.92 & 0.974 & \textbf{284.68} \\ 
         & 0.2 & 41.02 & 0.89 & 0.93 & \textbf{0.976} &  \\ 
         & 0.5 & 41.12 & 0.95 & 0.94 & 0.975 & \\
         & 0.99 & 41.09 & 1.16 & 0.93 & 0.974 &  \\
         & 0.9999 & \textbf{41.77} & 1.57 & \textbf{0.95} & 0.970 &  \\ \hline
        BCE & 0.001 & 36.15 & \textbf{0.06} & 0.70 & 0.732 & 309.90 \\ 
         & 0.2 & 39.96 & 0.11 & 0.94 & \textbf{0.996} &  \\ 
         & 0.5 & 40.05 & 0.20 & 0.96 & 0.995 & \\
         & 0.99 & \textbf{41.09} & 0.62 & \textbf{1.00} & 0.984 &  \\
         & 0.9999 & 40.31 & 1.84 & \textbf{1.00} & 0.952 &  \\ \hline
         &  &  & 6 types of changes &  &  &  \\ \hline
        CPD & 0.001 & 38.98 & \textbf{0.83} & 0.81 & 0.931 & \textbf{317.41} \\ 
         & 0.7 & 40.13 & 1.13 & 0.86 & 0.958 &  \\ 
         & 0.9 & 40.20 & 1.14 & 0.87 & \textbf{0.959} &  \\ 
         & 0.99 & 40.42 & 1.25 & 0.88 & \textbf{0.959} &  \\ 
         & 0.9999 & \textbf{42.15} & 2.59 & \textbf{0.91} & 0.947 &  \\ \hline
        BCE & 0.001 & 12.29 & \textbf{0.00} & 0.16 & 0.382 & 356.14 \\ 
         & 0.7 & 40.23 & 0.40 & 0.95 & \textbf{0.984} &  \\ 
         & 0.9 & 40.53 & 0.70 & 0.97 & 0.982 &  \\ 
         & 0.999 & 41.42 & 2.98 & \textbf{0.99} & 0.931 &  \\ 
         & 0.9999 & \textbf{41.76} & 4.76 & 0.98 & 0.893 &  \\ \hline
         &  &  & 8 types of changes &  &  &  \\ \hline
        CPD & 0.001 & 37.20 & \textbf{1.07} & 0.78 & 0.903 & \textbf{327.02} \\ 
         & 0.5 & 38.37 & 1.48 & 0.83 & 0.929 &  \\ 
         & 0.9 & 39.41 & 1.71 & 0.84 & 0.929 &  \\ 
         & 0.999 & 41.89 & 2.69 & \textbf{0.88} & \textbf{0.937} &  \\ 
         & 0.9999 & \textbf{43.45} & 4.03 & \textbf{0.88} & 0.923 &  \\ \hline
        BCE & 0.001 & 16.41 & \textbf{0.00} & 0.17 & 0.576 & 350.65 \\ 
         & 0.5 & 39.68 & 0.27 & 0.88 & \textbf{0.974} &  \\ 
         & 0.9 & 41.00 & 1.31 & \textbf{0.95} & 0.973 &  \\ 
         & 0.999 & 43.53 & 5.78 & 0.94 & 0.891 &  \\ 
         & 0.9999 & \textbf{45.24} & 9.24 & 0.89 & 0.829 &  \\ \hline
         &  &  & 10 types of changes &  &  &  \\ \hline
        CPD & 0.001 & 39.07 & \textbf{0.99} & 0.81 & 0.940 & \textbf{296.34} \\ 
         & 0.2 & 39.71 & 1.27 & 0.83 & 0.943 &  \\ 
         & 0.7 & 40.14 & 1.45 & 0.83 & 0.944 &  \\ 
         & 0.999 & 40.68 & 1.68 & 0.86 & \textbf{0.950} &  \\ 
         & 0.9999 & \textbf{41.02} & 1.99 & \textbf{0.87} & 0.947 &  \\ \hline
        BCE & 0.001 & 21.21 & \textbf{0.00} & 0.25 & 0.660 & 338.22 \\ 
         & 0.2 & 40.27 & 0.22 & 0.92 & \textbf{0.986} &  \\ 
         & 0.7 & 41.39 & 0.87 & \textbf{0.97} & 0.981 &  \\ 
         & 0.999 & 42.96 & 6.78 & 0.93 & 0.871 &  \\ 
         & 0.9999 & \textbf{52.24} & 15.22 & 0.66 & 0.747 &  \\ \hline \\
\caption{Quality metrics for a selected set of thresholds. The \qm$\uparrow$\qm\ sign marks the metrics we would like to maximize, \qm$\downarrow$\qm\ -- to minimize.}
\label{table:metrcis}
\end{longtable}
%
\vspace{-1cm}

%

Based on the results obtained, the following conclusions can be drawn:
\begin{itemize}
\item For the smallest thresholds, the model with BCE loss tends to make many false alarms, resulting in low mean detection delay and low mean time to FA. At the same time, the model with principled CPD loss does not make false alarms and produces quality results according to these $2$ metrics and the Covering metric as well.
\item On the other hand, for the highest thresholds, the model with BCE loss has a large mean detection delay that is a sign of FN predictions. In this case, the model with CPD loss outperforms it in terms of Covering and $F_1$-score.
\item These two peculiarities prove that the models with custom CPD loss are more sure in their predictions than the ones with BCE loss: even a minor fluctuation of change probabilities is often a sign of the change point.
\item In the middle cases, both models behave similarly. The area under the detection curve is a metric of the model`s overall performance considering various alarm thresholds. The models with CPD loss have lower AUC metrics in all the experiments we conducted. Thus, they make better predictions on average.
\item Considering the main metrics, we can conclude that the more types of distribution changes we have, the worse the models' performances are. Both approaches struggle when the problem becomes more complex. 
\end{itemize}

\section{Acknowledgements}

The work of Alexey Zaytsev was supported by the Russian Foundation for Basic Research grant  20-01-00203. 
The work of Evgenya Romanenkova was supported  by the Russian Science Foundation (project 20-71-10135). 

\section{Conclusion}

We investigated the problem of change point detection in the case of multiple data distributions for methods based on representation learning and estimated the efficiency of the considered approach in terms of the main metrics of change point detection quality. 

We confirm that the models can work well for such data, but performance worsens while the number of distribution types increases. We also show that the principled model is more confident in its predictions for all cases. Thus, it is easier to chose a threshold for predictions. 
\printbibliography[heading=bibintoc,]
\end{document}